\documentclass[10pt,twocolumn,letterpaper]{article}

\usepackage{authblk}
\usepackage{cvpr}

\usepackage{xcolor}
\usepackage{colortbl}
\usepackage{arydshln}
\usepackage{graphicx}
\usepackage{wrapfig}
\usepackage{booktabs}
\usepackage{multirow}
\usepackage{makecell}
\usepackage{amssymb}
\usepackage{caption}
\definecolor{cvprblue}{rgb}{0.21,0.49,0.74}
\usepackage[pagebackref,breaklinks,colorlinks,allcolors=cvprblue]{hyperref}
\colorlet{lightblue}{cvprblue!15}
\def\confName{CVPR}
\def\confYear{2027}

\title{CT-Navigator: Towards Interpretable and Generalizable CT Understanding through Agentic Perception and Tool-Invocation}

\renewcommand{\Affilfont}{\small}

\makeatletter

\ifdefined\AB@affillist
  \author[1,2,*]{Yibing Fu}
  \author[2,3,\protect\textdagger]{Huidong Xie}
\else
  \author[1,2]{Yibing Fu\textsuperscript{*}}
  \author[2,3]{Huidong Xie\textsuperscript{\textdagger}}
\fi

\author[3]{Yihao Niu}
\author[4]{Wei Ji}
\author[5]{Heming Dong}
\author[1]{Ziyue Wang}
\author[5]{Jingyu Chen}
\author[2]{Jun Zhao}
\author[2]{Haoyuan Hu}
\author[3]{Hongming Shan}
\author[5]{Lai Jiang}
\author[5]{Mai Xu}

\ifdefined\AB@affillist
  \author[1,\protect\textdagger]{Yueming Jin}
\else
  \author[1]{Yueming Jin\textsuperscript{\textdagger}}
\fi

\affil[1]{National University of Singapore}
\affil[2]{JD Health Inc.}
\affil[3]{Fudan University}
\affil[4]{Yale University}
\affil[5]{Beihang University}

\ifdefined\AB@affillist
  \renewcommand{\AB@affillist}{%
    \Affilfont
    \textsuperscript{1}National University of Singapore \quad
    \textsuperscript{2}JD Health Inc. \quad
    \textsuperscript{3}Fudan University\\
    \textsuperscript{4}Yale University \quad
    \textsuperscript{5}Beihang University%
  }
\fi

\makeatother

\begin{document}
\maketitle

\begingroup
\renewcommand{\thefootnote}{\fnsymbol{footnote}}
\footnotetext[2]{Corresponding authors: Huidong Xie and Yueming Jin.}
\footnotetext[1]{Work done during an internship at JD Health Inc.}
\endgroup

\begin{abstract}
Clinical computed tomography (CT) interpretation inherently requires progressive visual exploration, where radiologists navigate from whole scans to individual slices and regions of interest and interact with external tools when necessary. However, existing 3D medical multimodal large language models (MLLMs) typically connect a specialized 3D visual encoder to a language model, without fully leveraging the native 2D visual and agentic capabilities of generalist MLLMs. Trained on large-scale one-hop CT instruction-following data, these models also have limited ability to perform iterative reasoning and  explicitly ground their conclusions in visual evidence, limiting their support for traceable CT interpretation. To solve these problems, we propose \textbf{agentic perception} as an alternative paradigm inspired by radiologists' interpretation workflows. We introduce \textbf{CT-Navigator}, a framework that harnesses generalist MLLMs' 2D visual and agentic capabilities to progressively ground CT interpretation in explicit, traceable visual evidence through hierarchical visual exploration and radiology tool invocation. To build these capabilities, we curate 8,904 visual chain-of-thought (V-CoT) trajectories for cold-start agentic supervised fine-tuning. We then optimize the policy through reinforcement learning on 33,219 questions, with data selection and dynamic rollout sampling prioritizing informative learning signals. Finally, we introduce CTNavi-Bench, a 3D CT visual question answering benchmark comprising 7,000 questions spanning anatomical regions throughout the human body. CT-Navigator achieves 79.50\% accuracy on CTNavi-Bench, surpassing the strongest baseline by 17.59\%, and demonstrates strong generalization across three external CT benchmarks. The code, model, and benchmark will be released on \href{https://github.com/fyb99/ct_navigator}{GitHub}.
\end{abstract}
 
\section{Introduction}
\label{sec:intro}

\begin{figure}
\centering
\includegraphics[width=.9\linewidth]{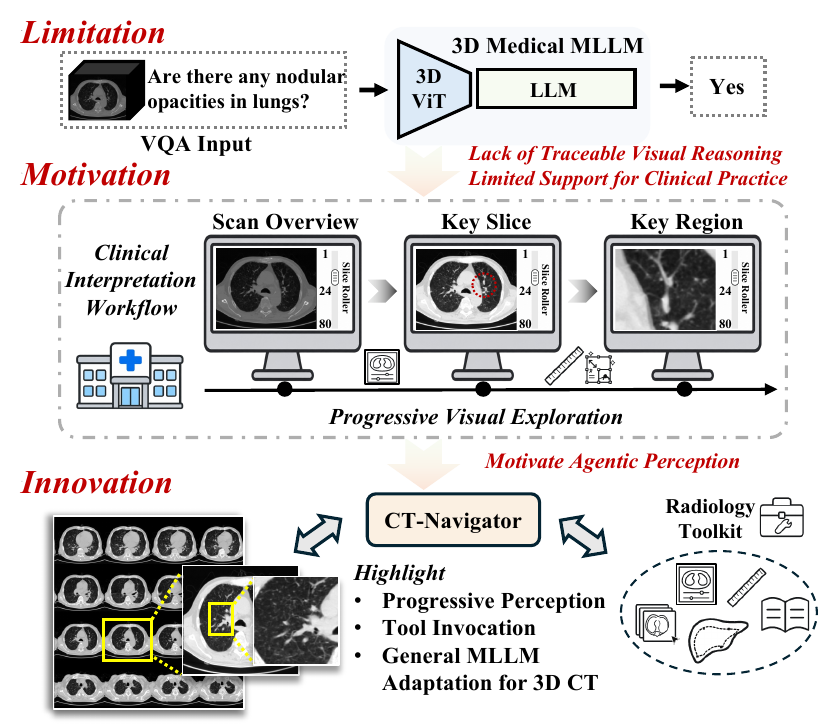}
\caption{\textbf{Motivation for CT-Navigator.} Existing 3D medical MLLMs typically couple a 3D visual encoder with a language model to generate answers directly from CT volumes, providing limited traceable visual reasoning. Inspired by the progressive visual navigation workflow of radiologists in clinical practice, we propose CT-Navigator, which adapts generalist MLLMs for CT understanding in an agentic perception setting.}
\label{fig1}
\end{figure}

Computed tomography (CT) is one of the most widely used non-invasive medical imaging modalities for clinical diagnosis, disease monitoring, and post-treatment follow-up~\citep{haydel2000indications}. It provides volumetric anatomical information that enables clinicians to inspect organs and surrounding structures within human bodies. In clinical practice, as illustrated in Fig.~\ref{fig1}, radiologists actively navigate CT volumes, progressing from a scan-level overview to key slices and clinically relevant regions of interest (ROIs)~\cite{drew2013informatics}. They also interact with radiology tools to localize anatomical structures and measure findings such as cysts and lung nodules. This iterative interpretation paradigm is significantly labour-intensive and time-consuming. As the number of CT examinations grows worldwide, radiologists face an increasingly heavy workload~\citep{bruls2020workload}, underscoring the importance of reliable AI-assisted systems to facilitate CT interpretation~\citep{oren2020artificial, topol2019high}.

Recent advances in medical multimodal large language models (MLLMs)~\citep{jiang2025hulu, xu2026lingshu, wang2026medagent} offer promising opportunities for AI-assisted CT interpretation. Several CT-oriented 3D MLLMs have been designed for tasks such as visual question answering (VQA) and radiology report generation~\citep{bai2024m3d, wu2025towards}. Most existing 3D MLLMs connect a specialized 3D visual encoder to a language model to accommodate volumetric inputs and perform large-scale training subsequently. In clinical practice, however, radiologists inspect 2D slices rather than directly view 3D volumes, combining visual exploration with external tool-interacting. This clinical workflow motivates a new paradigm for CT understanding in which generalist MLLMs progressively inspect 2D slices and local views rather than directly processing 3D volumes. Aligned with radiologists' clinical interpretation workflows, this paradigm explicitly connects diagnostic conclusions to supporting visual evidence, making the interpretation process explicit and traceable. More importantly, this paradigm can harness the native perception, planning, and tool-use capabilities of generalist 2D MLLMs, which benefit from large-scale multimodal pretraining and dedicated agentic training.


Based on this paradigm, we reformulate 3D CT understanding as \textbf{an agentic perception problem and adapt generalist MLLMs to actively explore CT slices and ground their interpretation in explicit, traceable visual evidence}. As illustrated in Fig.~\ref{fig1}, this formulation follows the progressive visual evidence accumulation process in clinical practice. We introduce \textbf{CT-Navigator}, an agentic framework for 3D CT interpretation through \textbf{hierarchical agentic perception}. CT-Navigator first examines a montage overview of sampled CT slices, then progressively identifies and inspects relevant slices and ROIs, guided by the clinical question and its evolving visual observations. We further equip CT-Navigator with a \textbf{radiology toolkit} to gather complementary evidence for CT interpretation. Throughout this process, CT-Navigator proactively decides where to inspect and how to acquire visual evidence before reaching a conclusion, producing traceable trajectories.

Conventional CT instruction-tuning datasets primarily consist of VQA pairs, providing solely answer-level supervision with limited guidance for visual evidence acquisition and tool invocation. To develop these agentic capabilities, we curate 8,904 high-quality visual chain-of-thought (V-CoT) trajectories combining human-annotated visual navigation with LLM-synthesized tool interactions. These trajectories provide explicit supervision for progressive visual perception and tool invocation, enabling efficient adaptation of generalist MLLMs through cold-start supervised fine-tuning (SFT). We then optimize the model policy through agentic reinforcement learning (RL) to learn generalizable strategies for CT interpretation. To support efficient RL training, we introduce a dynamic data sampling strategy that prioritizes questions that provide informative feedback for policy improvement.

Finally, since most existing 3D CT benchmarks focus on specific anatomical regions, we introduce \textbf{CTNavi-Bench}, a clinical finding grounded benchmark comprising 7,000 questions across anatomical regions throughout the body. Experiments on CTNavi-Bench and three external CT benchmarks demonstrate that CT-Navigator outperforms existing MLLMs and exhibits strong out-of-distribution generalization. Further ablation studies show consistent gains of 20\% over the base model across 4 benchmarks, demonstrating the effectiveness of the agentic perception paradigm and our training strategy. Our technical contributions are summarized as follows:

\begin{itemize}
\item \textbf{Hierarchical Agentic Perception.}
We reformulate 3D CT understanding as an agentic perception problem and introduce CT-Navigator, which adapts generalist MLLMs through hierarchical exploration of key slices and ROIs, along with radiology tool invocation.

\item \textbf{V-CoT Curation \& Data-efficient Training.}
We curate 8,904 agentic V-CoT trajectories to provide process-level supervision for visual navigation and tool invocation. We develop a two-stage training strategy combining cold-start SFT with RL, using dynamic data sampling during RL to support efficient policy optimization.

\item \textbf{Benchmark Construction.}
We construct CTNavi-Bench to evaluate CT understanding across anatomical regions and levels of granularity.

\item \textbf{Experiments.}
Comprehensive experiments on CTNavi-Bench and three external CT benchmarks demonstrate superior performance and strong out-of-distribution generalization of CT-Navigator.
\end{itemize}





\section{Related Work}
\subsection{Medical MLLMs}

Vision–language learning aims to integrate visual and textual information into unified representations for multimodal understanding. Early approaches primarily explored multimodal alignment between image features and text features~\citep{wang2018tienet, fu2026unleashing}. Subsequently, CLIP-style~\cite{radford2021learning} methods adopted contrastive pretraining on large-scale medical image-caption corpus~\citep{lin2023pmc,zhang2023biomedclip}, facilitating transferable visual representations. With the recent advancement of MLLMs~\citep{liu2023visual, alayrac2022flamingo}, substantial progress has been made in building medical MLLMs~\citep{li2023llava,tu2024towards}, which support a
broad range of tasks such as VQA and radiology report generation. However, these models are primarily designed for 2D medical images and offer limited support for interpreting volumetric scans. To narrow this gap, several works have explored extending MLLMs to 3D medical imaging domain. For example, RadFM~\citep{wu2025towards} incorporates both 2D and 3D image–text pairs into radiological pretraining. M3D-LaMed~\citep{bai2024m3d} brings retrieval, question answering, and segmentation into a unified 3D MLLM framework. More recently, OmniCT~\citep{lin2026omnict} strengthens spatial representations across slices and volumes using spatial-consistency constraints and organ-level alignment. Photon~\citep{fang2026photon} focuses on efficient volumetric processing by visual token pruning. Despite these advances, existing methods rely heavily on large-scale instruction tuning, which may encourage memorization of response patterns rather than explicitly ground their responses and reasoning in relevant visual evidence.

\begin{table}[t]
\centering
\caption{Comparison of existing 3D CT VQA benchmarks in terms of train/test samples, anatomy coverage and V-CoT annotations.}
\label{tab:ct_vqa_benchmarks}

\resizebox{\columnwidth}{!}{%
\begin{tabular}{ccccc}
\toprule
Benchmark & Train & Test & Coverage & V-CoT \\
\cmidrule(lr){1-1}
\cmidrule(lr){2-5}
M3D-VQA~\cite{bai2024m3d}
& $\sim$ 482k & 13,792 & Whole-body
& $\times$  \\
RadGenome-Chest CT~\citep{zhang2025development}
& $\sim$ 1.37M & 86,189 & Chest
& $\times$  \\
3D-RAD~\citep{gai20263d}
& $\sim$ 136k & 33,910 & Chest
& $\times$ \\
DeepTumorVQA~\citep{chen2025vision}
& $\sim$ 356k & 39,650 & Abdomen
& $\times$ \\
\rowcolor{lightblue} Our CTNavi-Bench
& 42,123 & 7,000 & Whole-body
& $\checkmark$ \\
\bottomrule
\end{tabular}%
}

\end{table}

\subsection{3D CT Understanding Benchmark}

The development of 3D MLLMs has prompted increasing efforts to benchmark 3D medical understanding. As summarized in Table~\ref{tab:ct_vqa_benchmarks}, M3D-Bench~\citep{bai2024m3d} covers a broad range of tasks, including image–text retrieval, VQA, and segmentation. 3D-RAD~\citep{gai20263d} focuses on CT-based diagnostic reasoning, with questions addressing abnormality recognition and longitudinal assessment. DeepTumorVQA~\citep{chen2025vision} introduces a large-scale benchmark with 29 medical tasks of
4 dimensions for abdominal CT interpretation. Nevertheless, two limitations remain within these established benchmarks. \textit{First, template-based question generation from a limited set of CT volumes may introduce content redundancy and overrepresent tasks that do not reflect clinical practice.} For example, DeepTumorVQA includes quantitative questions derived from segmentation masks, such as HU value ranges, which may place disproportionate emphasis on measurements that are not routinely documented in radiology reports. \textit{Second, existing benchmarks primarily provide large collections of repetitive VQA pairs but lack interpretable V-CoT annotations to guide visual reasoning.} These answer-level annotations provide limited guidance on how to progressively acquire relevant evidence, potentially encouraging memorization of response patterns rather than active visual reasoning.

\section{Benchmark}
\label{sec:method}
\subsection{Benchmark Establishment}
As summarized in Table~\ref{tab:ct_vqa_benchmarks}, most existing 3D medical VQA benchmarks focus on specific anatomical regions. To address this limitation, we construct CTNavi-Bench for evaluating CT understanding across multiple anatomical regions.

\noindent\textbf{Data Curation}: For data collection, we collaborated with a top-tier medical institution to collect CT examinations from real-world clinical visits, which provide scans covering different anatomical regions with their associated radiology reports. After filtering low-quality data, we collected 14,853 CT scans from 11,215 patients in total. During the data partitioning stage, we strictly followed a patient-level protocol to avoid patient-level data leakage, resulting in 12,011 and  2,842 scans for building training and evaluation corpus. Our work is conducted according to the principles of the Declaration of Helsinki and all the data are de-identified.

\begin{figure*}
\centering
\includegraphics[width=.95\linewidth]{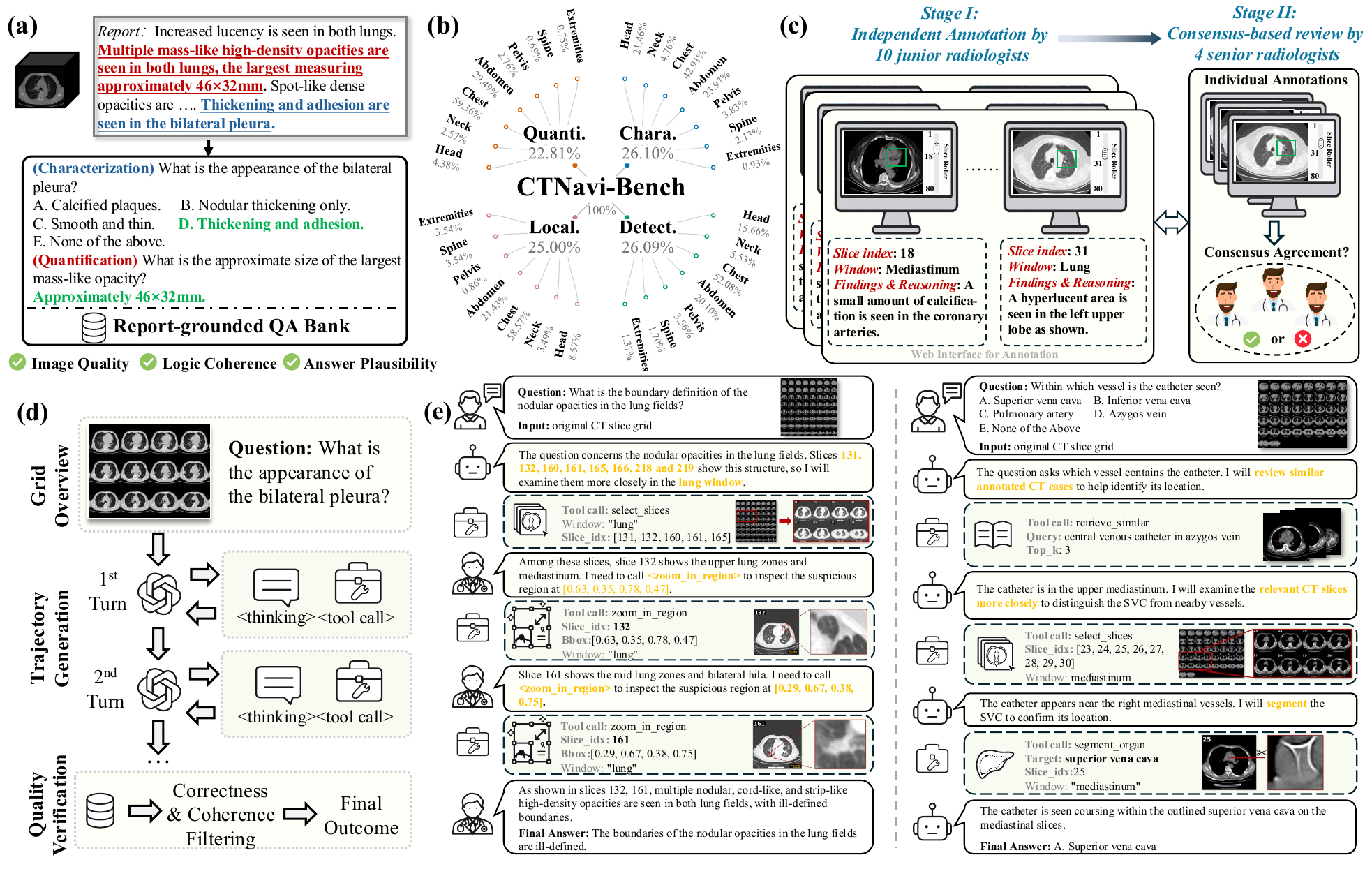}
\caption{\textbf{Overview of CTNavi-Bench construction and agentic trajectory curation.} (a) Pipeline for report-grounded QA generation and quality control. (b) Question distribution by category and anatomical region. (c) Pipeline for visual navigation annotation by radiologists. (d) Pipeline for LLM-based agentic trajectory synthesis. (e) Example trajectories showing visual navigation and tool interactions.}
\label{fig:data}
\end{figure*}

\noindent\textbf{Benchmark Construction}: Furthermore, we construct VQA samples from the radiology reports associated with volumetric scans. The construction pipeline includes data-quality filtering, QA generation, and balanced sampling. The overall pipeline is illustrated in Fig.~\ref{fig:data}-(a). Specifically, we initially exclude scans that fail to meet predefined quality criteria such as scans with very few slices or unparsable reports. Then, we employ GPT-5.6-Sol to extract clinically meaningful findings from each report and transform them into multiple-choice (MCQ) or open-ended questions. To build comprehensive assessment dimensions, we prompt the LLM to generate questions spanning four complementary aspects, from global perception to quantitative measurement. The four question categories include:
\begin{itemize}
    \item Characterization: Questions examine qualitative attributes of identified findings.
    \item Detection: Questions assess the presence of findings or identify abnormalities within an anatomical region.
    \item Localization: Questions assess the location of identified findings such as laterality or organ compartments.
    \item Quantification: Questions require explicit counting or measurement of anatomical structures and lesions.
\end{itemize}

\label{sec:vqa_gen}

Finally, human experts are involved to review the clarity and clinical correctness after this generation step. We then apply selective sampling to balance question categories and clinical content for maximizing question diversity. Meanwhile, we balance answer options in MCQ and maintain approximately equal proportions of MCQ and open-ended questions in final benchmark. Ultimately, we sample a compact set of 7,000 VQA pairs that preserves broad diversity and clinical relevance.

\subsection{CoT Annotation}

The training corpora in existing 3D CT VQA benchmarks primarily consist of VQA pairs, without explicit supervision on how to acquire relevant visual evidence. This outcome-oriented supervision alone provides limited guidance for traceable clinical reasoning. Therefore, we curate 8,904 agentic V-CoTs from the training corpus of CTNavi-Bench that integrate visual exploration and tool interactions.

We collect radiologist annotations about visual exploration through a web-based CT viewer that supports scrolling through slices, adjusting window settings, and zooming into ROI. As shown in Fig.\ref{fig:data}-(c), annotation proceeds in two stages. In the first stage, 10 junior radiologists independently review CT scans to inspect the abnormalities documented in original report. For each reported finding, they select key slices, choose appropriate window settings, and draw bounding boxes around the relevant regions. Each annotation records the slice index, window setting, bounding box coordinates, and accompanying findings and reasoning descriptions. In the second stage, four senior radiologists jointly review the individual annotations and retain only cases on which they reach consensus. We then link each question to the reviewed annotations through its referenced finding and use these annotations to guide V-CoT generation. Specifically, starting from a slice montage, each trajectory proceeds to select relevant slices with appropriate window settings, inspect marked regions through necessary zooming operations, and reason from the gathered visual evidence to answer the question.

Beyond visual navigation and CT slice inspection, MLLMs can use external tools such as retrieval and quantitative measurement tools to obtain additional evidence. We generate trajectories that supervise this capability using GPT-5.6-Sol. Specifically, GPT-5.6-Sol receives a question and an initial CT slice montage with access to the radiology toolkit described in Sec.~\ref{sec:tool}. Along the reasoning process, the model reasons over the available evidence to decide whether and how to invoke a tool. The returned output informs subsequent reasoning and tool selection until the model produces a final answer. We retain only coherent trajectories that include at least one tool invocation and produce a correct answer at the end. MCQ questions are evaluated by matching the predicted and reference options, while open-ended questions are evaluated by Qwen3.5-397B-A17B~\cite{qwen3.5} for semantic consistency with the reference answers. In total, we collect 8,904 V-CoT trajectories. Radiologist-annotated trajectories supervise progressive visual perception, while GPT-generated trajectories complement this supervision by illustrating how to acquire and integrate evidence through external tools.

\begin{figure*}
\centering
\includegraphics[width=.9\linewidth]{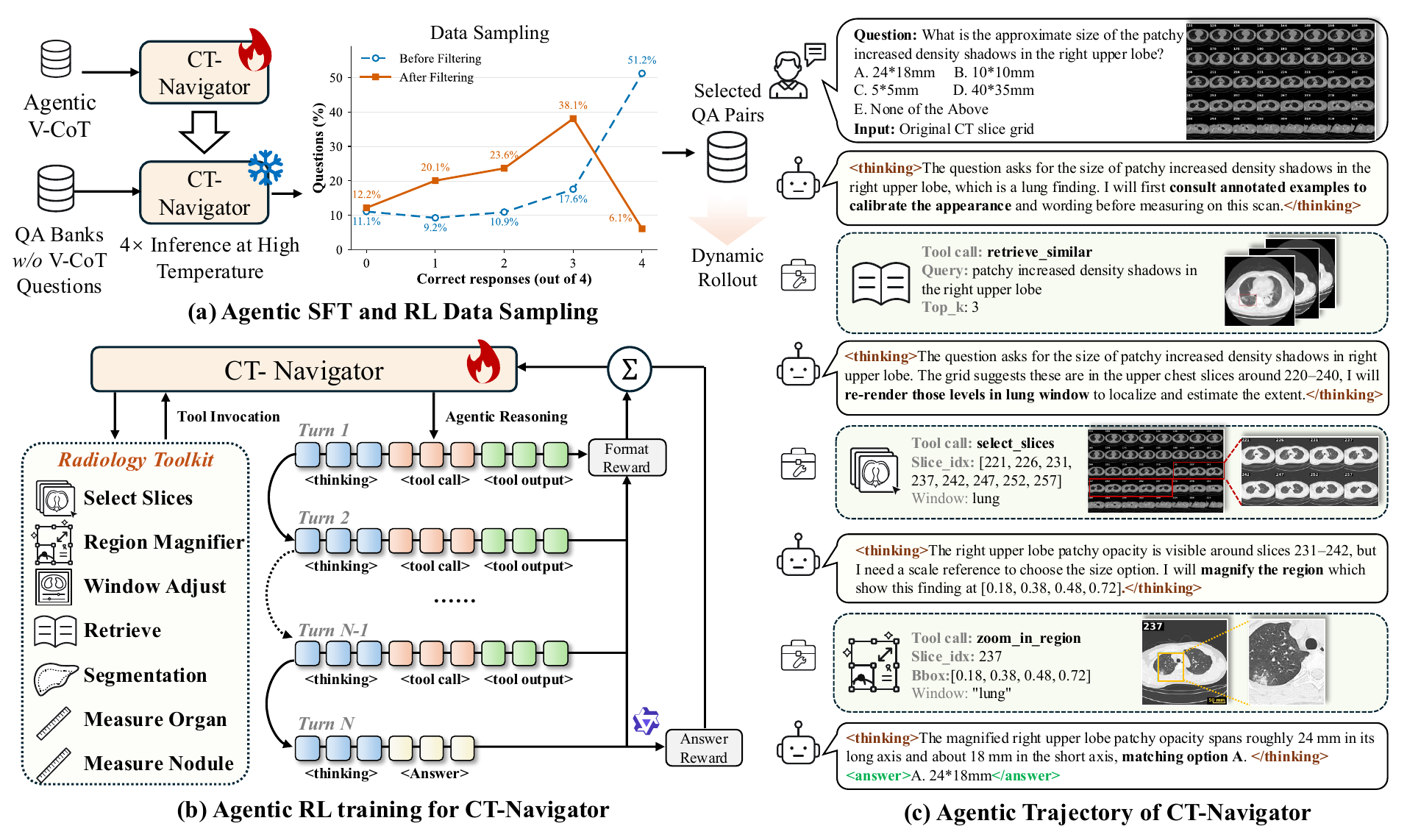}
\caption{\textbf{Overview of CT-Navigator}. (a) Cold-start SFT on V-CoT trajectories and RL data sampling guided by four high-temperature rollouts per question. (b) Agentic RL training with multi-turn tool interactions, optimized using a hybrid reward. (c) An example trajectory generated from CT-Navigator interleaving reasoning with multiple tools before producing the final answer.}
\label{fig:method}
\end{figure*}

\section{Methodology}
\subsection{Framework Overview}
We reformulate 3D CT understanding as an agentic perception problem. Specifically, given a question $q$ and a CT volume $V$, CT-Navigator initially receives $q$ and a montage image $M$ consisting of 64 uniformly sampled slices from $V$. The montage provides a global overview in a 2D format accessible to general-purpose MLLMs. Guided by the question and its evolving observations, CT-Navigator progressively examines relevant clips, slices, and ROIs, while invoking specialized tools available in the radiology toolkit to gather complementary evidence.

Following previous agentic MLLMs~\cite{Chng_2026_CVPR,qin2026tool,qin2026surgo}, CT-Navigator follows a ReAct loop~\cite{yao2022react} that interleaves reasoning, actions, and observations. Let $h_0=(q,M)$ denote the initial context and $h_{t-1}$ denote the interaction history before step $t$. The model policy $\pi_\theta$ generates a reasoning step $r_t$ and an action $a_t$ conditioned on this history:
\begin{equation}
    (r_t, a_t) \sim \pi_\theta(\cdot \mid h_{t-1}),
    \qquad
    a_t \in \mathcal{A}_{\mathrm{tool}} \cup \mathcal{A}_{\mathrm{answer}},
\end{equation}
where $\mathcal{A}_{\mathrm{tool}}$ contains the available tool calls and $\mathcal{A}_{\mathrm{answer}}$ denotes final-answer actions. For a tool call $a_t=(\mathcal{T}_t,u_t)$, the system executes $\mathcal{T}_t$ with arguments $u_t$ on the input context $\hat{V}$ and appends the returned observation $o_t$ to the history:
\begin{equation}
    o_t = \mathcal{T}_{t}(\hat{V};u_t),
    \qquad
    h_t = (h_{t-1},r_t,a_t,o_t).
\end{equation}
The updated history guides subsequent reasoning and action selection until the model selects an action in $\mathcal{A}_{\mathrm{answer}}$ and synthesizes the accumulated evidence into its final answer.
\subsection{Radiology Toolkit}
\label{sec:tool}
The initial montage provides global anatomical context but may omit details needed for further inspection. We equip CT-Navigator with six radiology tools for visual navigation and complementary evidence acquisition. The model selects tools based on the question and its history context and uses these tool outputs to inform further inspection or produce its final answer.

\noindent\textbf{Slice selection}.
The model requests one or more slice indices from the original input montage overview to inspect a suspected finding. These may include slices absent from the initial montage. The tool assembles the requested slices into a new montage with each slice displayed at $512 \times 512$ pixels. This supports progressive localization from a global overview to key slices that best show the finding.

\noindent\textbf{Window display adjustment}.
In clinical practice, CT windowing adjusts the displayed attenuation range to emphasize different tissues and findings. CT-Navigator can choose among five window presets for brain, chest, mediastinum, abdomen, and bone. We incorporate the window setting as an argument to visual tools such as slice selection to reduce interaction turns and token overhead. The requested slices are rendered under the chosen window before being returned for inspection.

\noindent\textbf{Region magnifier}. The region magnifier enables closer inspection of a finding within a CT slice. Following the zoom-in operation in DeepEyes~\cite{hong2026deepeyesv}, the model specifies the region using a bounding box $[x_1, y_1, x_2, y_2]$. The tool returns two complementary views in one call: the original slice with the box highlighted and an enlarged crop of the selected region. The crop allows detailed inspection of the finding's shape and margins, while the marked slice retains its location relative to global anatomy.

\noindent\textbf{Anatomy segmentation}.
The segmentation tool uses VISTA3D~\cite{he2025vista3d} to delineate anatomical structures requested by the model. It returns the original CT image with the segmented region highlighted and a separate ROI view. The highlighted image identifies the target within the surrounding anatomy, while the ROI allows the model to inspect its shape and boundaries in detail.

\noindent\textbf{Cross-modal retrieval}. Retrieval-based~\cite{lewis2020retrieval, Chng_2026_CVPR} tools can enhance MLLM reasoning by supplying relevant external knowledge and exploit their strong in-context learning capabilities. Motivated by this, we equip CT-Navigator with a cross-modal retrieval tool that provides annotated CT slice examples as references during interpretation. We construct the retrieval corpus from representative slices in our training set and additional examples from DeepLesion~\cite{yan2018deeplesion}. Each entry $(I_i,b_i,c_i)$ contains a CT slice, a bounding box marking the finding, and a descriptive caption. CT-Navigator formulates a text query $q_{\mathrm{r}}$ based on its current observations. We encode the query and corpus captions as $\mathbf{e}_{\mathrm{r}}=E(q_{\mathrm{r}})$ and $\mathbf{e}_i=E(c_i)$ using Qwen3-Embedding-0.6B~\cite{zhang2025qwen3}. We rank entries by cosine similarity $s_i=\mathbf{e}_{\mathrm{r}}^{\top}\mathbf{e}_i/(\|\mathbf{e}_{\mathrm{r}}\|_2\|\mathbf{e}_i\|_2)$ and return the three highest-ranked slices with their findings highlighted by a rectangle and captions attached.

\noindent\textbf{Measurement}.
The measurement toolkit provides quantitative evidence for assessing organ size, tissue attenuation, and pulmonary nodule dimensions. For organ measurement, the model specifies the target organs and the tool applies VISTA3D~\cite{he2025vista3d} to obtain their segmentation masks. It then computes each organ’s volume, length, width, and attenuation in Hounsfield units (HU). Additionally, we deploy a dedicated detector to localize pulmonary nodules for measuring their sizes. The detector is an nnDetection model~\cite{baumgartner2021nndetection} trained on public datasets ~\cite{wang2025duke,setio2017validation,peeters2026benchmarking}, with its encoder initialized from nnFoundation~\cite{harsy2026nnfoundation}. It predicts candidate bounding boxes in chest CT. We use a VISTA3D lung mask to filter out candidates outside the lungs, then return the estimated physical sizes of the retained nodules.

\subsection{Learning Objective}
We train CT-Navigator in two stages: cold-start SFT followed by agentic RL.

\noindent\textbf{Cold-start SFT.}
We first fine-tune the model on the curated V-CoT dataset $\mathcal{D}_{\mathrm{SFT}}$ to establish foundational capabilities in hierarchical visual perception and tool invocation. At step $t$ of trajectory $i$, the history $h_{i,t-1}$ contains the question, initial montage, and preceding interactions. We supervise the target model response $y_{i,t}$ by minimizing
\begin{equation}
    \mathcal{L}_{\mathrm{SFT}}(\theta)
    = -\mathbb{E}_{(h_{i,t-1},\,y_{i,t}) \sim \mathcal{D}_{\mathrm{SFT}}}
    \left[\log \pi_\theta(y_{i,t} \mid h_{i,t-1})\right].
\end{equation}

\noindent\textbf{RL data sampling.}
To improve RL efficiency, we prioritize questions that provide informative feedback for policy optimization. The candidate pool consists of 72,256 training questions not used for SFT. As illustrated in Fig.~\ref{fig:method}-(a), the SFT model generates four trajectories per question at a high sampling temperature. We group questions by the number of correct answers and adjust their sampling proportions. Questions answered correctly in all four attempts are substantially downsampled because they offer little variation in answer correctness. Finally, this selection yields 33,219 questions for RL training.

\noindent\textbf{Agentic RL.}
We optimize the policy using GSPO~\cite{zheng2025groupsequencepolicyoptimization} through multi-turn interactions with the radiology toolkit, as shown in Fig.~\ref{fig:method}-(b). During training, we adopt the dynamic sampling strategy from DAPO~\cite{yu2025dapoopensourcellmreinforcement} to prioritize rollout groups with nonzero reward variance, ensuring informative feedback for policy updates. For a trajectory $\tau$, the total reward is $R(\tau)=R_{\mathrm{acc}}(\tau) + 0.2\times R_{\mathrm{format}}(\tau)$. The format reward checks compliance with the interaction protocol. The performance reward compares the final answer against the reference through option matching for MCQ and an LLM judge for open-ended questions. Fig.~\ref{fig:method}-(c) illustrates a trajectory that gathers evidence through successive tool calls before answering.
\begin{table*}[t]
\centering
\caption{Comparison results on CTNavi-bench.
Multiple-choice questions are evaluated using accuracy (\%).
Open-ended questions are evaluated using LLM-judged accuracy
(\%) and raw BERTScore-$\mathrm{F}_1$ (BSF).
Avg. denotes overall accuracy.}
\label{tab:model_comparison}

\resizebox{.87\textwidth}{!}{%
\begin{tabular}{l*{13}{c}}
\toprule
\multirow{3}{*}{\textbf{Model}}
& \multicolumn{4}{c}{\textbf{Multiple-choice Questions}}
& \multicolumn{8}{c}{\textbf{Open-ended Questions}}
& \multirow{3}{*}{\textbf{Avg.}} \\
\cmidrule(lr){2-5}
\cmidrule(lr){6-13}
& Chara. & Det. & Loc. & Quan.
& \multicolumn{2}{c}{Chara.}
& \multicolumn{2}{c}{Det.}
& \multicolumn{2}{c}{Loc.}
& \multicolumn{2}{c}{Quan.}
& \\
\cmidrule(lr){6-7}
\cmidrule(lr){8-9}
\cmidrule(lr){10-11}
\cmidrule(lr){12-13}
& Acc. & Acc. & Acc. & Acc.
& Acc. & BSF
& Acc. & BSF
& Acc. & BSF
& Acc. & BSF
& \\
\midrule

\multicolumn{14}{c}{\textit{Closed-source Generalist MLLMs}} \\

GPT-5.6-Luna
& 72.73 & 81.82 & 63.66 & 49.87
& 68.82 & 89.96
& 78.42 & 93.17
& 34.29 & 90.29
& 18.77 & 89.16
& 59.44 \\

GPT-5.6-Sol
& 73.49 & 81.27 & 68.34 & 53.88
& 72.98 & 90.79
& 85.76 & 93.68
& 34.74 & 89.53
& 17.40 & 89.75
& 61.91 \\


\midrule

\multicolumn{14}{c}{\textit{Open-source Generalist MLLMs}} \\

InternVL3-8B
& 48.85 & 64.07 & 49.83 & 38.97
& 47.05 & 89.51
& 54.55 & 93.08
& 25.26 & \textbf{92.85}
& 2.75 & 90.55
& 42.13 \\

Qwen3-VL-8B
& 53.23 & 55.42 & 48.11 & 46.99
& 43.54 & 88.46
& 50.82 & 83.34
& 18.63 & 88.07
& 7.38 & 89.67
& 41.03 \\

Qwen3.5-9B
& 51.37 & 72.62 & 49.26 & 44.61
& 49.67 & 83.27
& 70.32 & 82.49
& 12.00 & 85.12
& 13.02 & 87.44
& 46.06 \\

Qwen3.6-27B
& 61.66 & 59.26 & 43.09 & 47.99
& 47.48 & 83.58
& 54.00 & 82.59
& 18.17 & 85.70
& 21.03 & 87.15
& 44.54 \\

Qwen3.8-27B
& 69.44 & 61.56 & 51.31 & 48.25
& 51.53 & 84.88
& 59.80 & 82.92
& 19.66 & 86.14
& 22.28 & 88.02
& 48.53 \\
\midrule

\multicolumn{14}{c}{\textit{Medical MLLMs}} \\

Hulu-Med-7B
& 70.21 & 73.06 & 61.83 & 52.88
& 57.99 & 92.85
& 81.60 & 96.54
& 20.11 & 88.23
& 21.40 & \textbf{91.52}
& 55.61 \\

Hulu-Med-32B
& 77.88 & 77.77 & 63.89 & 54.89
& 58.64 & 90.82
& 80.18 & 95.62
& 20.69 & 87.53
& 24.66 & 88.38
& 58.06 \\

Lingshu-7B
& 61.66 & 61.88 & 54.97 & 47.99
& 44.31 & 87.37
& 63.53 & 88.28
& 15.09 & 87.20
& 14.89 & 90.22
& 46.11 \\

Lingshu-32B
& 69.66 & 63.31 & 58.06 & 52.76
& 50.55 & 87.71
& 60.46 & 88.30
& 17.60 & 88.41
& 13.39 & 89.65
& 48.83 \\

HealthGPT-8B
& 72.07 & 59.58 & 63.09 & 54.51
& 51.53 & 85.34
& 73.82 & 88.53
& 14.29 & 85.73
& 17.40 & 87.18
& 51.40 \\
\midrule

\multicolumn{14}{c}{\textit{3D Medical MLLMs}} \\

M3D-Phi-3-4B
& 56.63 & 67.58 & 53.03 & 33.21
& 49.56 & 89.14
& 62.98 & 89.48
& 19.31 & 89.86
& 7.01 & 90.79
& 44.51 \\

M3D-Llama-2-7B
& 37.79 & 62.98 & 42.40 & 29.20
& 39.93 & 86.84
& 66.16 & 83.17
& 15.09 & 89.80
& 3.88 & 88.88
& 37.94 \\

RadFM-14B
& 26.83 & 51.48 & 21.94 & 24.31
& 46.83 & 92.42
& 64.95 & 94.45
& 17.71 & 92.26
& 7.63 & 91.50
& 33.40 \\

Merlin
& 40.96 & 52.57 & 30.51 & 15.41
& 35.89 & 91.32
& 56.74 & 92.66
& 19.43 & 92.56
& 10.51 & 88.86
& 33.49 \\

OmniCT-7B
& 71.74 & 64.18 & 58.63 & 45.74
& 54.92 & 87.56
& 72.95 & 89.48
& 16.23 & 85.72
& 22.40 & 87.44
& 51.54 \\

Photon-3B
& 57.17 & 67.25 & 53.37 & 51.38
& 42.78 & 88.68
& 67.25 & 88.27
& 10.63 & 87.73
& 7.26 & 88.55
& 45.27 \\

\rowcolor{lightblue}
\textbf{CT-Navigator}
& \textbf{90.47} & \textbf{96.28}
& \textbf{79.09} & \textbf{75.94}
& \textbf{81.62} & \textbf{95.55}
& \textbf{94.85} & \textbf{97.22}
& \textbf{69.83} & 92.05
& \textbf{42.43} & 88.96
& \textbf{79.50} \\
\bottomrule
\end{tabular}%
}

\end{table*}

\section{Experiments}
\subsection{Implementation Details}

We initialize CT-Navigator from Qwen3.5-9B~\cite{qwen3.5} and run cold-start SFT on 8 NVIDIA H200 GPUs for 1 epoch using a learning rate of $1 \times 10^{-6}$ and a global effective batch size of 32. For RL, we allocate 40 H200 GPUs for training, with 16 for rollout generation and 24 for policy updates. We train the model using a learning rate of $1 \times 10^{-6}$, a global effective batch size of 384 and a rollout group size of 4.

\subsection{Main Results}

\noindent\textbf{Results on CTNavi-Bench.}
Table~\ref{tab:model_comparison} compares our CT-Navigator with 18 generalist, medical, and 3D medical MLLMs on CTNavi-Bench. Generalist baselines include GPT-5.6-Luna and GPT-5.6-Sol, InternVL3~\citep{zhu2025internvl3}, and the Qwen series~\citep{yang2025qwen3, team2026qwen3, qwen3.6-27b}. We also evaluate medical MLLMs including Hulu-Med~\citep{jiang2025hulu}, Lingshu~\citep{xu2026lingshu}, and HealthGPT~\citep{lin2025healthgpt}, alongside 3D medical MLLMs including M3D~\citep{bai2024m3d}, RadFM~\citep{wu2025towards}, Merlin~\citep{blankemeier2026merlin}, OmniCT~\citep{lin2026omnict}, and Photon~\citep{fang2026photon}. As summarized in Table~\ref{tab:model_comparison}, our model achieves an overall accuracy of 79.50\%, outperforming the strongest baseline GPT-5.6-Sol by 17.59\%. It achieves the highest semantic accuracy in all four question categories. In particular, open-ended localization and quantification accuracies reach 69.83\% and 42.43\%, exceeding the best competing results by 35.09\% and 17.77\%, respectively.

\noindent\textbf{Results on External Benchmarks.}
Table~\ref{tab:external_benchmarks} evaluates CT-Navigator on 3 external benchmarks including Merlin~\cite{blankemeier2026merlin}, CT-RATE~\cite{hamamci2026generalist}, and DeepTumorVQA~\cite{chen2025vision} beyond CTNavi-Bench in a zero-shot setting. Note that we use GPT-5.6-Sol for building QA pairs in Merlin and CT-RATE from their official released scans and reports. Following prior work~\citep{wang20263dmedagent}, we construct balanced evaluation subsets of 4,000 questions per benchmark by reducing redundancy and answer-position bias. As summarized in Table~\ref{tab:external_benchmarks}, our model achieves overall accuracies of 60.35\%, 68.88\%, and 59.88\%, exceeding the strongest competing results by 4.60\%, 3.48\%, and 7.40\% without any training on these benchmarks. For DeepTumorVQA, this comparison excludes Photon, whose dataset-specific fine-tuning results are reported as a specialist reference. These consistent gains across three external subsets demonstrate the model's strong generalization ability beyond CTNavi-Bench.

\subsection{Ablation Study}

\begin{table*}[t]
\centering
\caption{Comparison results on Merlin, CT-RATE, and DeepTumorVQA.
Results are reported as accuracy (\%).
Char., Recog.,  Loc., Meas., Vis.R. and Med.R.  denote characterization,
recognition, localization, measurement, visual and medical reasoning.}
\label{tab:external_benchmarks}
\resizebox{.95\textwidth}{!}{%
\begin{tabular}{l*{15}{c}}
\toprule
\multirow{2}{*}{\textbf{Model}}
& \multicolumn{5}{c}{\textbf{Merlin}}
& \multicolumn{5}{c}{\textbf{CT-RATE}}
& \multicolumn{5}{c}{\textbf{DeepTumorVQA}} \\
\cmidrule(lr){2-6}
\cmidrule(lr){7-11}
\cmidrule(lr){12-16}
& Char. & Recog. & Loc. & Meas. & \textbf{Avg.}
& Char. & Recog. & Loc. & Meas. & \textbf{Avg.}
& Recog. & Meas. & Vis.R. & Med.R. & \textbf{Avg.} \\
\midrule

\multicolumn{16}{c}{\textit{Closed-source Generalist MLLMs}} \\

GPT-5.6-Luna
& 54.50 & 77.80 & 57.60 & 33.10 & 55.75
& \textbf{72.90} & 84.90 & 59.10 & 44.70 & 65.40
& 71.84 & 40.27 & 39.99 & 24.94 & 49.30 \\

GPT-5.6-Sol
& 47.00 & 71.30 & \textbf{59.90} & 39.00 & 54.30
& 64.20 & 79.20 & \textbf{61.30} & 41.20 & 61.48
& 73.17 & 32.06 & 40.50 & 39.51 & 49.58 \\


\midrule

\multicolumn{16}{c}{\textit{Open-source Generalist MLLMs}} \\

InternVL3-8B
& 30.90 & 65.10 & 42.40 & 22.50 & 40.23
& 44.10 & 67.70 & 37.10 & 20.30 & 42.30
& 83.52 & 29.92 & 39.32 & 41.23 & 52.38 \\

Qwen3-VL-8B
& 33.70 & 57.50 & 45.30 & 35.00 & 42.88
& 50.50 & 57.20 & 41.60 & 39.00 & 47.08
& 64.01 & 30.60 & 39.02 & 37.53 & 45.45 \\

Qwen3.5-9B
& 30.90 & 57.00 & 32.80 & 23.80 & 36.13
& 53.90 & 63.70 & 43.00 & 28.10 & 47.18
& 44.49 & 29.92 & 38.58 & 41.23 & 38.93 \\

Qwen3.6-27B
& 36.50 & 56.40 & 42.70 & 32.90 & 42.13
& 57.80 & 65.30 & 44.00 & 35.60 & 50.68
& 83.96 & 22.61 & 32.59 & 38.02 & 48.30 \\

Qwen3.8-27B
& 44.50 & 56.10 & 44.70 & 22.90 & 42.05
& 66.20 & 62.50 & 48.70 & 26.50 & 50.98
& 76.20 & 30.37 & 41.98 & 42.47 & 51.03 \\

\midrule
\multicolumn{16}{c}{\textit{Medical MLLMs}} \\

Hulu-Med-7B
& 57.60 & 72.40 & 48.50 & 24.50 & 50.75
& 58.50 & 84.80 & 50.00 & 27.20 & 55.13
& 84.85 & 27.56 & 33.11 & 30.86 & 49.15 \\

Hulu-Med-32B
& 55.80 & 73.60 & 45.80 & 37.20 & 53.10
& 64.80 & 80.50 & 52.80 & 47.90 & 61.50
& 84.63 & 30.60 & 37.99 & 29.14 & 51.23 \\

Lingshu-7B
& 48.30 & 61.40 & 45.60 & 27.60 & 45.73
& 56.20 & 62.60 & 45.70 & 31.40 & 48.98
& \textbf{85.29} & 31.27 & 40.35 & 29.88 & 52.48 \\

Lingshu-32B
& 47.70 & 61.30 & 45.10 & 31.50 & 46.40
& 61.60 & 64.30 & 51.20 & 37.40 & 53.63
& 71.18 & 39.15 & 40.58 & 27.90 & 49.33 \\

HealthGPT-8B
& 48.90 & 59.60 & 47.80 & 18.70 & 43.75
& 55.60 & 69.00 & 56.80 & 26.80 & 52.05
& 82.26 & 27.11 & 34.07 & 32.10 & 48.63 \\

\midrule
\multicolumn{16}{c}{\textit{3D Medical MLLMs}} \\

M3D-Phi-3-4B
& 44.00 & 76.70 & 44.00 & 18.50 & 45.80
& 42.00 & 75.10 & 41.60 & 16.80 & 43.88
& 50.63 & 25.98 & 30.52 & 37.53 & 37.03 \\

M3D-Llama-2-7B
& 36.60 & 43.90 & 38.40 & 21.40 & 35.08
& 27.20 & 56.00 & 37.70 & 16.10 & 34.25
& 8.87 & 22.16 & 27.57 & 35.80 & 20.88 \\

RadFM-14B
& 24.70 & 52.20 & 23.80 & 23.80 & 31.13
& 25.00 & 52.20 & 22.70 & 25.00 & 31.23
& 42.57 & 24.52 & 30.38 & 37.28 & 33.90 \\

Merlin
& 28.80 & 49.90 & 32.10 & 19.30 & 32.53
& 32.80 & 49.20 & 28.40 & 17.90 & 32.08
& 49.08 & 22.95 & 29.64 & 35.80 & 35.35 \\

OmniCT-7B
& 55.70 & 64.70 & 46.00 & 18.40 & 46.20
& 55.90 & 72.20 & 48.70 & 19.00 & 48.95
& 62.97 & 24.07 & 30.60 & 30.62 & 40.10 \\

Photon-3B$^{\dagger}$
& 45.90 & 55.20 & 41.60 & \textbf{41.20} & 45.98
& 48.10 & 61.70 & 42.20 & 29.50 & 45.38
& \textcolor{gray}{83.15}
& \textcolor{gray}{73.34}
& \textcolor{gray}{65.48}
& \textcolor{gray}{61.23}
& \textcolor{gray}{72.78} \\

\rowcolor{lightblue}
\textbf{CT-Navigator}
& \textbf{67.00} & \textbf{81.00} & 52.50
& 40.90 & \textbf{60.35}
& 71.10 & \textbf{89.40} & 58.60
& \textbf{56.40} & \textbf{68.88}
& 79.31 & \textbf{53.77} & \textbf{52.25}
& 33.83 & \textbf{59.88} \\

\bottomrule
\end{tabular}%
}
\begin{minipage}{\textwidth}
\footnotesize
$^{\dagger}$Photon is specially fine-tuned on
DeepTumorVQA. These results are shown in gray for reference and excluded
from best-result highlighting.
\end{minipage}
\end{table*}

\noindent\textbf{Effect of Training Stages.}
We conduct ablation study on our training stages. As shown in Table~\ref{tab:training_stage_ablation}, we compare the base model (Qwen3.5-9B), agentic SFT, and subsequent RL training outcomes.
On CTNavi-Bench, agentic SFT increases overall accuracy from 46.06\% to 66.39\%, and RL further improves it to 79.50\%.
The same progression can be observed across the three external benchmarks, where RL improves over agentic SFT by 4.97\%, 5.23\%, and 6.28\% on Merlin, CT-RATE, and DeepTumorVQA, respectively.
The final model outperforms the base model by 33.44\% on CTNavi-Bench and 20.95\%–24.22\% across the external benchmarks. These consistent gains demonstrate the model's effectiveness in CT interpretation and its ability to generalize across datasets.

\noindent\textbf{Tool Configuration and Invocation Patterns.} We further investigate how different tool configurations affect CT understanding performance and how tool usage varies across question types.
Given the substantial computational cost of RL, we conduct tool configuration ablations at the agentic SFT stage. Table~\ref{tab:tool_ablation} compares variants whose training trajectories consist of different tool combinations. Adding retrieval to perception tools improves mean accuracy from 56.29\% to 57.84\%, and further adding measurement tools raises it to 59.75\%. The largest gain occurs on DeepTumorVQA, where accuracy increases from 43.70\% with perception tools alone to 53.60\% with the full toolkit. Perception with retrieval performs best on Merlin and CT-RATE. This may reflect differences in task composition, as DeepTumorVQA contains many organ measurement questions that benefit from measurement tools. Across all four benchmarks, the full toolkit achieves the highest mean accuracy.

\begin{figure}
\centering
\includegraphics[width=.87\linewidth]{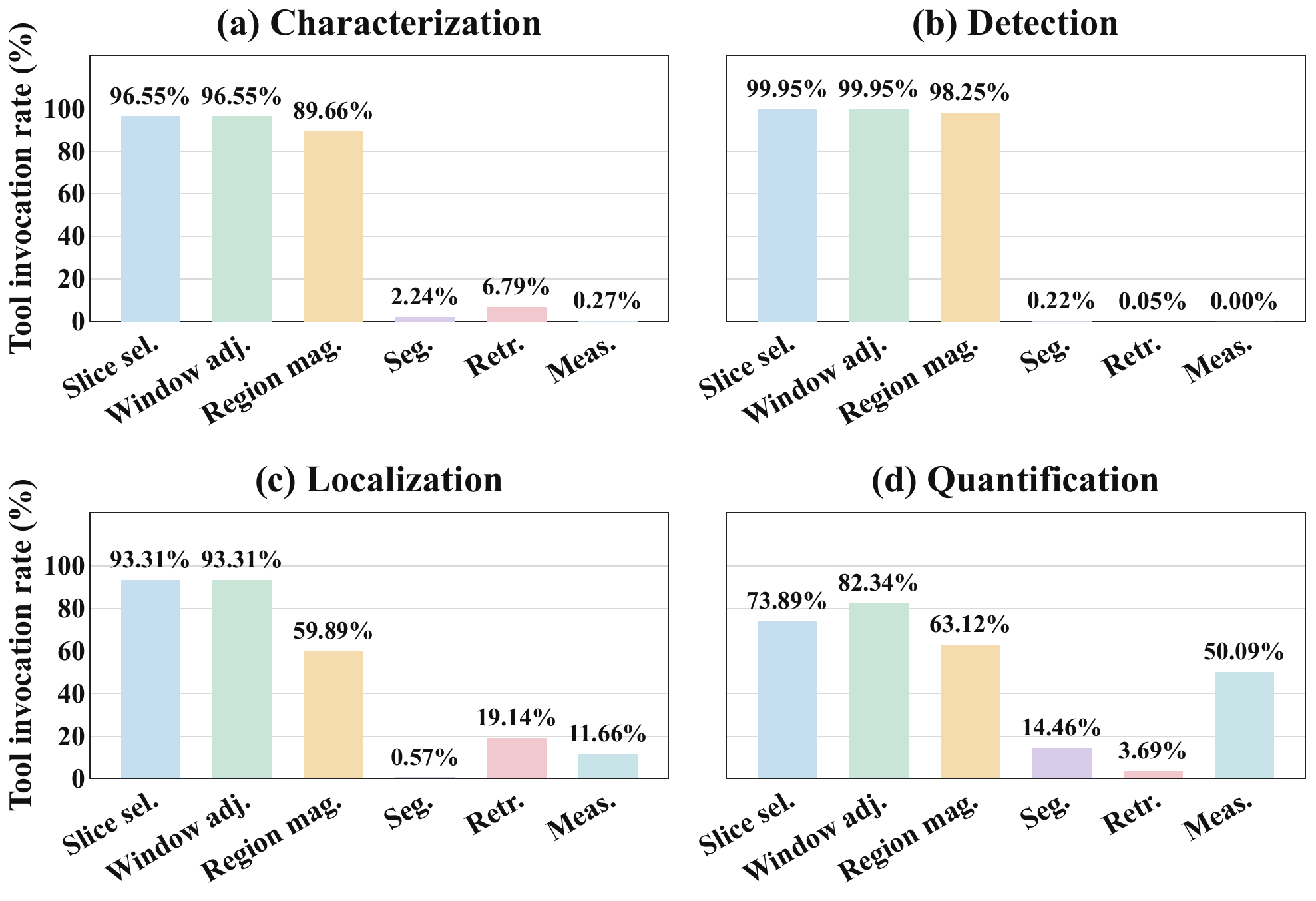}
\vspace{-1em}
\caption{Tool invocation rate across tasks on CTNavi-Bench.}
\label{fig:toolrate}
\end{figure}

Figure~\ref{fig:toolrate} further analyzes question-level tool invocation rates on CTNavi-Bench. Notably, slice selection, window adjustment, and region magnification are frequently invoked across question categories. These tools let the model actively inspect CT images at different locations, scales, and display settings, forming the foundation of progressive visual perception. Specialized tools show more selective usage. Retrieval is most frequent for localization at 19.14\%, whereas measurement and segmentation are most frequent for quantification at 50.09\% and 14.46\%, respectively. These patterns reflect task-dependent evidence acquisition.

\begin{table}[t]
\centering
\caption{Ablation study of training stages.
Gains denote the improvement of the final RL model
over base model.}
\label{tab:training_stage_ablation}
\resizebox{\columnwidth}{!}{%
\begin{tabular}{ccc llll}
\toprule
\textbf{\textit{Base}} & \textbf{\textit{SFT}} & \textbf{\textit{RL}}
& \textbf{Ours} & \textbf{Merlin}
& \textbf{RATE} & \textbf{DeepT} \\
\cmidrule(lr){1-3}
\cmidrule(lr){4-7}

$\checkmark$ & &
& 46.06 & 36.13 & 47.18 & 38.93 \\

$\checkmark$ & $\checkmark$ &
& 66.39 & 55.38 & 63.65 & 53.60 \\

\rowcolor{lightblue}
$\checkmark$ & $\checkmark$ & $\checkmark$
& \textbf{79.50} & \textbf{60.35}
& \textbf{68.88} & \textbf{59.88} \\

\cmidrule(lr){1-7}
\multicolumn{3}{c}{\textit{Gains} ($\Delta$)}

& \footnotesize + $33.44_\uparrow$ & \footnotesize + $24.22_\uparrow $
& \footnotesize + $21.70_\uparrow $ & \footnotesize + $20.95_\uparrow $ \\
\bottomrule
\end{tabular}}
\end{table}


\begin{table}[t]
\centering
\caption{Tool ablation results of agentic SFT. Percep., Retri. and Mea. denote the visual perception, retrieval and measurement tools, respectively.}
\label{tab:tool_ablation}

\resizebox{\columnwidth}{!}{%
\begin{tabular}{ccc ccccc}
\toprule
\textbf{\textit{Percep.}}
& \textbf{\textit{Retri.}}
& \textbf{\textit{Mea.}}
& \textbf{Ours}
& \textbf{Merlin}
& \textbf{RATE}
& \textbf{DeepT}
& \textbf{Avg} \\
\cmidrule(lr){1-3}
\cmidrule(lr){4-8}

$\checkmark$ & & 
& 65.01 & 55.35 & 61.10 & 43.70 & 56.29 \\

$\checkmark$ & $\checkmark$ &
& 64.29 & \textbf{55.78} & \textbf{64.55}
& 46.75 & 57.84 \\

\rowcolor{lightblue}
$\checkmark$ & $\checkmark$ & $\checkmark$
& \textbf{66.39} & 55.38 & 63.65
& \textbf{53.60} & \textbf{59.75} \\

\bottomrule
\end{tabular}%
}
\end{table}

\section{Conclusion}
In this paper, we introduced CT-Navigator, an agentic framework that adapts generalist MLLMs to 3D CT interpretation through hierarchical visual exploration and radiology tool invocation. To develop these capabilities, we curated 8,904 V-CoT trajectories and adopted a two-stage training strategy combining cold-start SFT with RL. Data selection and dynamic rollout sampling prioritize informative learning signals for policy improvement. Experiments on our proposed CTNavi-Bench and three external benchmarks demonstrate superior performance and generalization abilities of CT-Navigator.
In the future, we will scale up the training data to further improve the performance.

\section*{LLM Usage Statement}
In this work, we used ChatGPT (GPT-5.6-Sol) and OpenAI Codex to assist with synthetic data generation, method implementation, and manuscript preparation. The conceptual framework, scientific decisions, and interpretation of experimental results were determined by the authors. We reviewed all AI-assisted outputs and take full responsibility for the final submission, including its text, code, scientific claims, citation accuracy, experimental results, and reproducibility.

{
    \small
    \bibliographystyle{ieeenat_fullname}
    \bibliography{main}
}


\end{document}